\documentclass[12pt]{article}

\usepackage[letterpaper,margin=1in]{geometry}
\usepackage{scicite}
\usepackage{amsmath,amssymb} %
\usepackage{newtxtext,newtxmath} %
\usepackage{graphicx} 
\usepackage{epsfig}
\usepackage{makecell}
\usepackage[normalem]{ulem}  %
\usepackage{helvet}
\usepackage{caption}
\newcommand{\m}[1]{\mathrm{M}_{#1}}

\date{}
\renewenvironment{abstract}
    {\quotation}
    {\endquotation}

\makeatletter
\renewcommand{\fnum@figure}{\textbf{Figure \thefigure}}
\renewcommand{\fnum@table}{\textbf{Table \thetable}}
\makeatother

\newcommand{\name}{Pneu-ron }

\def\scititle{
Pneumatic neurons for soft robots enable inflate-and-fire networks for rhythmic motion
}
\title{\bfseries \boldmath \scititle}
\author{
	Dongting Li$^{1}$,
	Michael T. Tolley$^{1\ast}$,
	Nick G. Gravish$^{1\ast}$\and
	\small$^{1}$MAE Department, UC San Diego, La Jolla, 92093, US\and
	\small$^\ast$Corresponding author. Email: \{tolley;ngravish\}@ucsd.edu
}

\begin{document} 

\maketitle

\begin{abstract}
Animals coordinate their movements through distributed neural circuits, but soft robots still typically depend on external, centralized electronics for control.
Building soft robots that operate without centralized electronic controllers while remaining responsive to their environment remains a frontier challenge in soft robotics.
In this work we introduce a soft-robot control architecture inspired by leaky integrate-and-fire models of biological neural circuits.
The \textbf{Pneu}matic neu\textbf{ron} (Pneu-ron) is a soft actuator that unifies energy conversion, logic, and actuation in one component.
Each module combines a low-boiling-point fluid (LBF), a heater, and a mechanical switch into a self-excitable unit.
Boiling the LBF inflates the module and triggers excitation and inhibition of adjacent modules in a process we call ``inflate-and-fire''. 
When interconnected into excitatory–inhibitory rings, Pneu-rons generate stable, sequential oscillations whose frequency emerges from the material dynamics and environmental conditions.
By harnessing the inflation of Pneu-rons for actuation these networks can drive oscillatory locomotion of soft robots. 
Pneu-ron networks sustain oscillation under mechanical load and thermal variations, adapting through material physics rather than computation.
Dynamical modeling of these networks reveals a dimensionless bifurcation diagram that dictates the network's oscillatory behavior. 
Encoding logic and actuation into material-level modules presents a new modular architecture for adaptive, electronics controller-free, soft robots. 
\end{abstract}

Soft robots present novel opportunities for exploration, human-robot interaction, and inherently safe operation~\cite{rus_design_2015}. 
Yet, deployment of soft robots is challenged by current control approaches which rely on centralized digital control systems and bulky fluidic pumps. 
Animals on the other hand, move through complex environments with an agility that arises not from a central processor alone, but from the interaction of computational and mechanical elements distributed throughout the body~\cite{chiel_brain_1997}.
In invertebrates such as annelids and arthropods, rhythmic locomotion arises from central pattern generators (CPGs) distributed segmentally along the nerve cord rather than from a single pacemaker~\cite{marder_invertebrate_2005,kristan_neuronal_2005,mulloney_neurobiology_2012}.
These excitable networks couple tightly to local muscle and body mechanics, producing rhythms that tune to mechanical load and recover after disturbance~\cite{bassler_pattern_1998,grillner_biological_2006}.
Timing and coordination are negotiated between the neural circuit and the physical environment, yielding robust autonomy without high-level control~\cite{ijspeert_central_2008,ijspeert_swimming_2007,owaki_simple_2013}.
Achieving similar emergent coordination between the mechanical and control systems would present significant opportunities for the operation of soft robotics. 

Recent work in fluidic autonomy has begun to close this gap.
Soft pneumatic ring oscillators produce electronics-free walking gaits from a single constant-pressure source~\cite{preston_soft_2019,drotman_electronics-free_2021}.
Soft logic gates supply computational building blocks~\cite{preston_digital_2019,picella_pneumatic_2024,conrad_3d-printed_2024}.
Physically synchronized self-oscillating limbs reach speeds orders of magnitude beyond earlier soft robots~\cite{comoretto_physical_2025}.
Modular soft oscillators can also coordinate through strain, producing collective patterns across loosely coupled units~\cite{ceron_soft_2021}.
Such autonomous rhythm is not limited to pneumatic actuation.
Chemical reaction-driven self-oscillating gels generate peristaltic waves within a polymer network~\cite{yoshida_self-oscillating_1996, maeda_peristaltic_2008,yashin_pattern_2006, proskurkin_experimental_2020}, and responsive-material oscillators swim under constant illumination~\cite{zhao_soft_2019}.
More recently, bistable mechanical lattices have been shown to propagate signals with no working fluid at all~\cite{raney_stable_2016,yasuda_mechanical_2021}.
Together, these systems establish that material dynamics can generate and coordinate rhythmic behavior without centralized electronic control.

\begin{figure}[h!]
 \centering
 \includegraphics[width=1\linewidth]{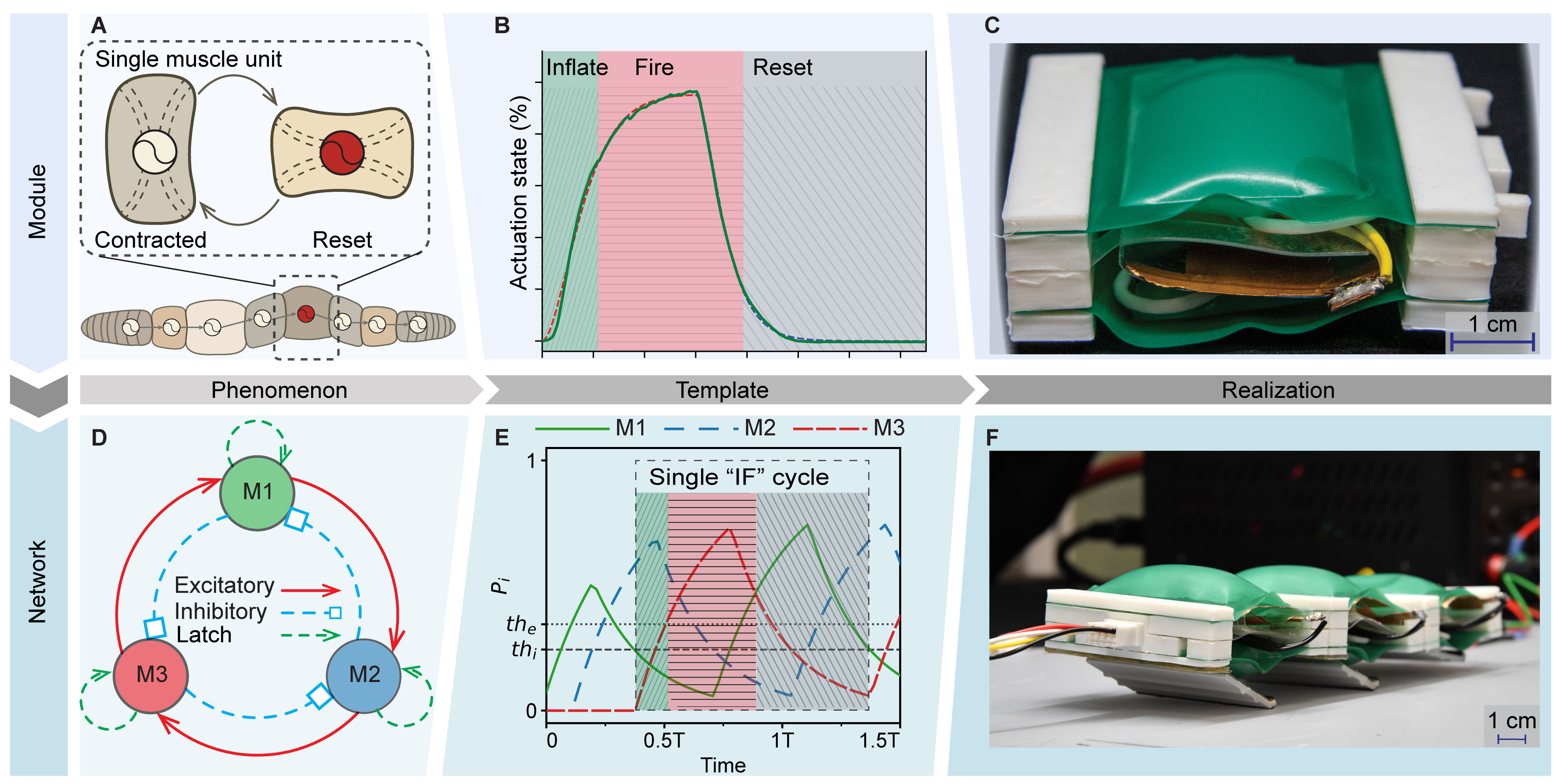}
 \caption{
 \textbf{From biological peristalsis to soft excitable rings.}
 (A) Contraction-relaxation of a single segment of a worm;
 (B) Pressure curve of \name during activation following the inflate-and-fire template;
 (C) Single \name module;
 (D) 3-segment CPG built upon excitatory-inhibitory connections;
 (E) Simulated actuation state of 3-segment CPG using inflate-and-fire template over 1.5 times of its period (T);
 (F) Connected \name ring for locomotion.
 }
 \label{fig:1_inspiration}
\end{figure}

Prior embodiments of self-oscillating robots  are typically fabricated as a single object and not inherently modular.
The oscillatory dynamics and the coupling between units are fixed by the geometry of the device rather than by ambient conditions~\cite{rothemund_soft_2018,laake_fluidic_2022,drotman_electronics-free_2021}, so obtaining a different gait usually requires building a different device.
Segmental biological central pattern generators are not restricted in this way, because the pattern is generated by the coupling between units rather than by the units themselves~\cite{marder_central_2001,grillner_biological_2006}.
One set of units can therefore produce several gaits, depending on how they are connected.

Furthermore, soft robotic oscillators typically require an external pressure source such as an onboard pump or a tethered supply line. 
Such pneumatic circuits are capable of controller-free rhythm, but changing the network topology generally requires rerouting or rebuilding the fluidic circuit, and mobile robots must carry along a pump for actuation.
It would be desirable to compartmentalize actuation within soft robots, enabling modularity and potentially relieving the need for pumps.
Generating the pressure inside each unit removes that shared network, so each unit becomes its own pressure source.
Neighboring units are therefore coupled electrically rather than through the exchange of working fluid, allowing the same units to be disconnected and rewired into different network topologies after fabrication. This capability could support devices in remote settings where electrical power is available but pneumatic supplies are impractical.

Inspired by the leaky integrate-and-fire (LIF) model of biological neurons and neural circuits~\cite{hodgkin_quantitative_1952, abbott_lapicques_1999,bean_action_2007, burkitt_review_2006}, we introduce the Pneu-ron, a soft module that realizes a similar cycle through thermodynamics and mechanical switching (Fig.~\ref{fig:1_inspiration}B).
Boiling a low-boiling-point fluid (LBF) inflates the module and eventually triggers a mechanical switch that excites and inhibits its neighbors.
Once the heating is removed, the pressure falls through passive cooling of the LBF.
This cycle of inflation, triggering, and cooling is what we call ``inflate-and-fire''.
The same pressure also serves two roles at once, acting as the internal state variable that sets the order of signaling and as the mechanical output that contracts the body.

Using the phase-change of LBF for actuation, however, limits how much force a module can produce.
A low-boiling-point fluid develops far less force than a pressurized supply line, too little to drive a snap-through valve directly.
Our modules therefore hold their state electrically, which separates excitation and inhibition onto two independent pressure thresholds.
The interval between those thresholds becomes a design parameter rather than a consequence of valve geometry.
When inflation reaches the mechanical switching threshold, the electrical connections within the module close a regenerative path, and the module resets as it dissipates heat after termination (Fig.~\ref{fig:1_inspiration}C).
A dual-threshold switch thereby provides excitatory, inhibitory, and self-latching pathways in one component (Fig.~\ref{fig:1_inspiration}C), serving as a building block for embodied intelligence.

Connected in a ring topology illustrated in Fig.~\ref{fig:1_inspiration}D, these modules form a compliant central pattern generator that can generate peristaltic locomotion of a soft robot without external electronic control (Fig.~\ref{fig:1_inspiration}F).
In this manuscript we demonstrate the design principles and capabilities of soft excitable Pneu-ron networks.
First, we show that sustained oscillation requires all three signaling pathways: excitation, inhibition, and self-latching.
Second, we predict its operating range with a dimensionless two-parameter bifurcation diagram and confirm the predicted boundaries under mechanical, thermal, and electrical perturbation.
Third, because coordination emerges from the topology rather than from a stored program, a severed ring can be rewired into a new gait.

\section{Results}

\subsection{Minimum signaling architecture for sustained network oscillation}
\label{sec:chain_propagation}

Modular CPGs in biology produce coordinated rhythmic activity through the interaction of locally excitable units, where timing and ordering are set by the connections between them.
We follow this principle and consider a network of identical leaky integrate-and-fire units, each modeled as a first-order Resistor-Capacitor (RC) element whose normalized internal pressure $P_i\in[0,1]$ rises while its heater is powered ($s_i=1$, $s_i$ stands for the state of activation signal) and decays otherwise,
\begin{equation}
\frac{dP_i}{dt}=\frac{s_i-P_i}{\tau(s_i)},
\qquad
\tau(s_i)=
\begin{cases}
\tau_{\mathrm{on}}, & s_i=1,\\
\tau_{\mathrm{off}}, & s_i=0,
\end{cases}
\label{eq:module_dynamics}
\end{equation}
and firing when $P_i$ crosses a threshold.
A single module is excitable but not bistable, meaning that it has one stable rest state and, once triggered, undergoes a transient pressure rise that decays back to that rest state.
We then identify what additional signals between such modules are required for a network of them to sustain a traveling wave, evaluating both open-chain and closed-ring topologies through first-order simulation.

\begin{figure}[h!]
 \centering
 \includegraphics[width=0.7\linewidth]{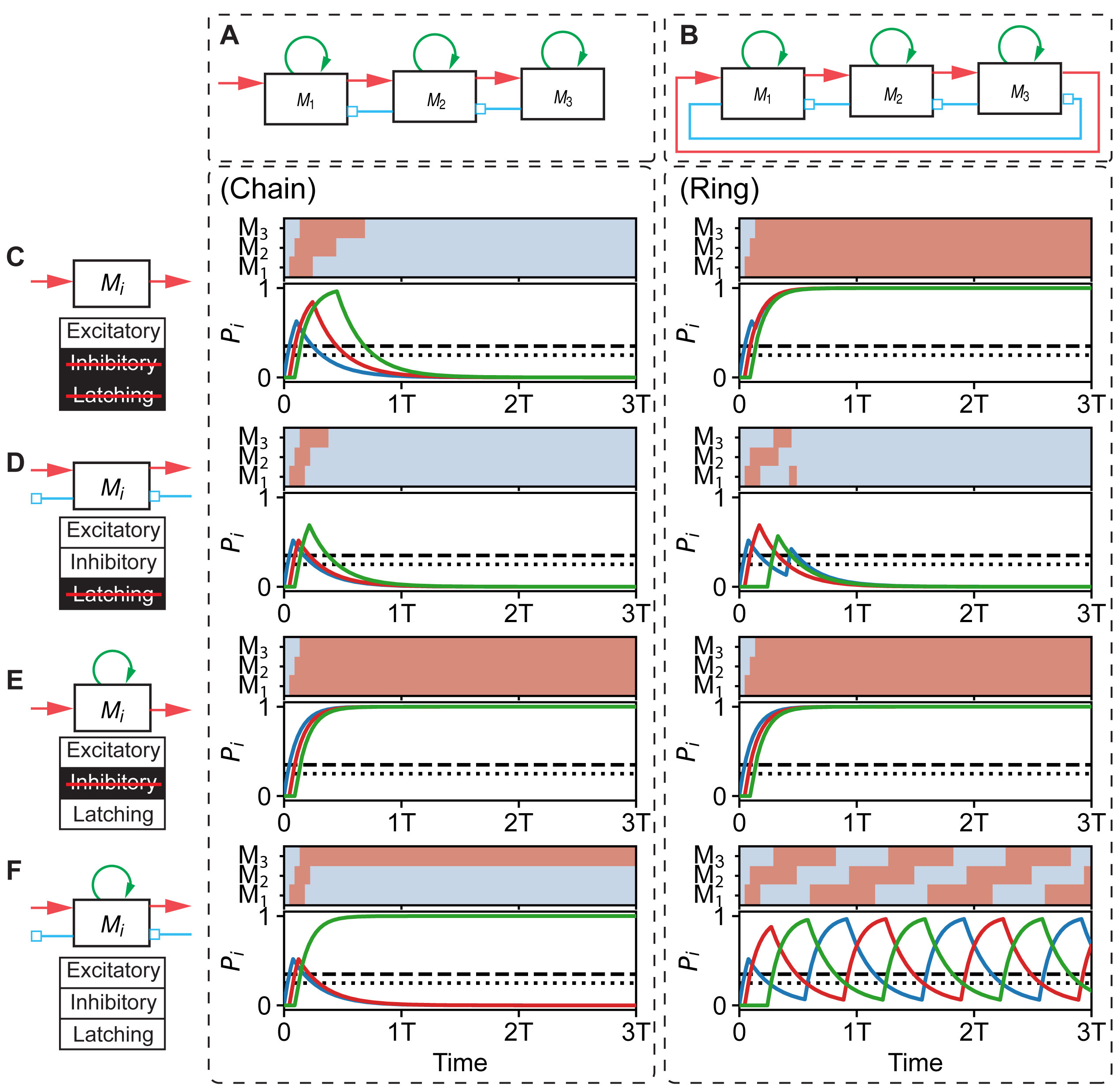}
 \caption{
 \textbf{Three signaling pathways required for stable oscillation.}
 (A) Open-chain configuration with no electrical connection from $\m{3}$ to $\m{1}$;
 (B) Ring architecture where the last module feeds back to the first, allowing returning waves;
 (C–F) Representative dynamical behaviors for different combinations of excitatory, inhibitory, and latching interactions. White background with black text means this signal pathway is on, while dark background means this channel is off, where ``T" denotes the period of oscillation.
 Sub-column (Chain) stands for chain configuration and (Ring) is ring configuration.
 }
 \label{fig:signal_speed}
\end{figure}

With only forward excitation between modules, a transient trigger applied to the first module ($\m{1}$; hereafter $\m{i}$ denotes the $i$-th module) initiates a transient wave that propagates a few modules but eventually fades.
This is because each module is an LIF unit with no internal memory. Once the triggering signal is removed, $\m{1}$ begins passive reset and ceases to provide excitation to the following modules. The chain settles to a uniform OFF state (Fig.~\ref{fig:signal_speed}C(chain)).
Closing the chain into a ring removes this dependency on the trigger but introduces saturation. Once $\m{1}$ activates $\m{2}$, the wave propagates around the ring, and $\m{3}$'s output re-energizes $\m{1}$.
Ring closure then forms a self-sustaining activation loop in which every module continuously excites its successor, and all modules saturate in the ON state (Fig.~\ref{fig:signal_speed}C(ring)).
Adding inhibition, in which a module cuts power to its predecessor upon crossing its inhibitory threshold, breaks the saturation but over-corrects.
The signal amplitude decays at each handoff in the chain configuration, and the ring collapses back to OFF (Fig.~\ref{fig:signal_speed}D).

We resolve this by adding an explicit self-latching pathway.
Once a module crosses its excitatory threshold, an internal positive-feedback path maintains its activation independently of continued upstream power.
We refer to this as logical bistability, in contrast to the mechanical bistability of snap-through valves and buckling membranes, where the two states are encoded in component geometry~\cite{rothemund_soft_2018,nemitz_soft_2020}.
Unlike a bistable mechanical switch, the state of our module is maintained by signal feedback through the network's electrical configuration.
With all three pathways present, the ring exhibits sustained traveling-wave oscillation (Fig.~\ref{fig:signal_speed}F(Ring)).
The wave front advances one module per handoff cycle, with one module rising toward its excitatory threshold while its predecessor is cooling past its inhibitory threshold.
Because a module stays conducting until its successor crosses the inhibitory threshold, and remains above ambient while it cools, a single traveling wave keeps exactly two adjacent modules warm.
We refer to this overlapping pair as the ``2-hot'' state, which is the minimal unit of the traveling wave.
The same architecture in chain configuration produces shift-register-like propagation, with the signal advancing to the terminal module which remains latched (Fig.~\ref{fig:signal_speed}F(Chain)).
In such networks of inflate-and-fire units with two signaling thresholds, the three pathways are not interchangeable, and only the full combination produces sequential, regenerative propagation.
Each pathway provides a structurally distinct function. Excitation provides forward signaling, inhibition provides upstream release, and latching provides module-level memory.

\subsection{Regeneration and clearance define the boundaries of the oscillatory region}
\label{sec:survival_map}

Having identified the minimal signaling architecture required for propagation in \S\ref{sec:chain_propagation}, we now determine the parameter conditions under which this architecture sustains oscillation.
Each of the two failure modes identified above (saturation and extinction) can be expressed as a constraint on the system's timing parameters. Together these constraints define the bounded region in which sustained oscillation is possible.

\begin{figure}[h!]
 \centering
 \includegraphics[width=1\linewidth]{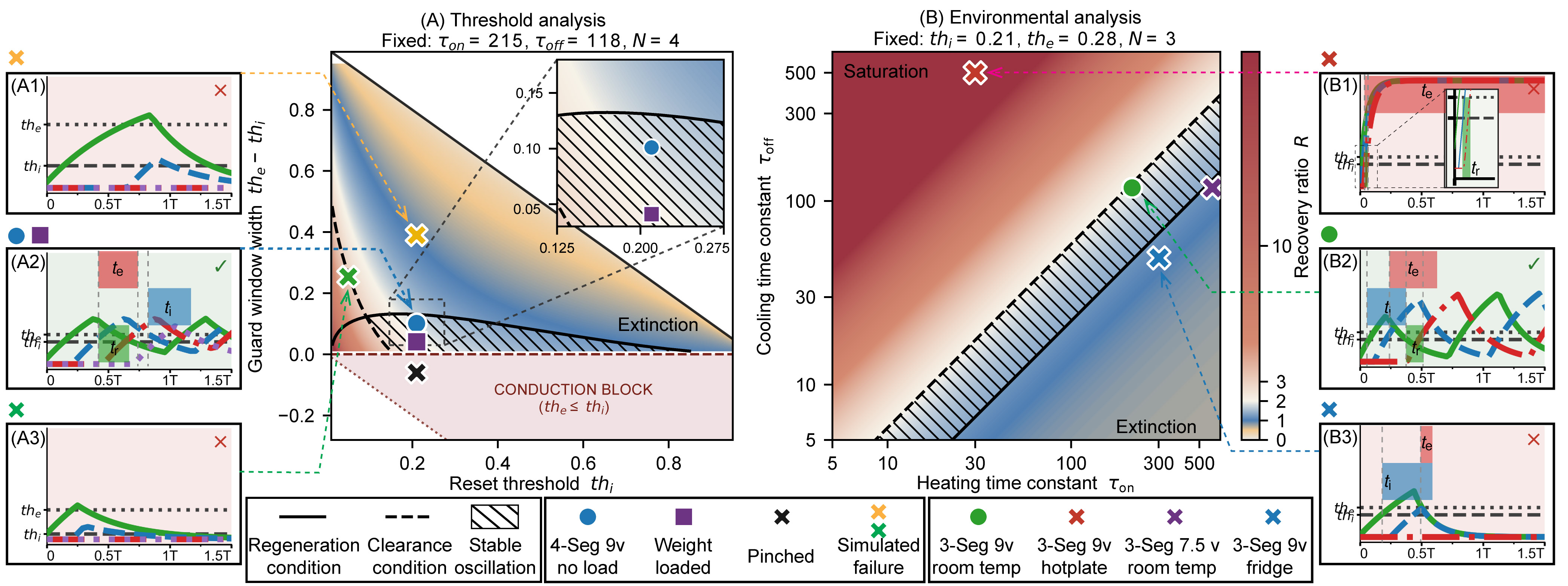}
 \caption{\textbf{Two-parameter bifurcation diagrams showing two representative cross-sections of the five-dimensional parameter space.}
 (A) Threshold-adjustment diagram. Insets (A1) and (A3) are per-module traces of two simulated failures (extinction); (A2) is the reference normal oscillation.
 (B) Environmental analysis based on heating and cooling time constants. Insets (B1)--(B3) show saturation, normal oscillation, and extinction.
 $R$ is shown as the map color and used as the primary parameter.
}
 \label{fig:survival_map}
\end{figure}

The first condition prevents extinction.
As shown in Fig.~\ref{fig:survival_map}B(3), propagation fails when inhibition surpasses upstream excitation before the downstream module can become self-latching, and the signal decays at each handoff. 
In this case, the excitation time $t_{\mathrm{e}}$ of $\m{2}$ (red block) is cut short by inhibition from the previous module before the rise time $t_{\mathrm{r}}$ of $\m{3}$ is complete, voiding the condition.
For successful transmission, the upstream module must remain active long enough for its successor to reach the excitatory threshold and self-latch.
This gives the first, \textit{regeneration} condition, expressed as a propagation gain $G$:
\begin{equation}
1 < t_{\mathrm{e}} / t_{\mathrm{r}} = G ,
\label{eq:regeneration_cond}
\end{equation}
where $t_{\mathrm{e}}$ is the duration for which module $M_{i-1}$ remains above its own excitatory threshold $th_e$, providing excitation to $M_i$ (red region in Fig.~\ref{fig:survival_map}A(2)), and $t_{\mathrm{r}}$ is the time required for $M_i$ to rise from its resting state to $th_e$ (green region).
When $t_{\mathrm{e}} < t_{\mathrm{r}}$, or when strong inhibition prevents $t_{\mathrm{r}}$ from being reached (Fig.~\ref{fig:survival_map}A(3)), regeneration fails and the system is incapable of generating steady-state oscillations. 
This condition applies to both open chains and closed rings.

The second condition prevents saturation and is specific to ring topologies as a \textit{clearance} condition.
As shown in Fig.~\ref{fig:survival_map}B(1), modules accumulate in the activated state when no mechanism resets them before the traveling wave returns.
Without inhibition, a module never shuts off. 
In this case, $t_{\mathrm{e}}$ never terminates for $\m{2}$ and $\m{3}$ has a short rising time $t_{\mathrm{r}}$.
For sustained oscillation, each module must recover below the excitatory threshold before being re-excited.

The recovery time available to a module is set by the wave's transit time around the ring, $(N-1)\,t_{\mathrm{r}}$, because the wave advances one module every $t_{\mathrm{r}}$.
The recovery time required is $t_{\mathrm{i}}$, the interval from the moment a module fires until its pressure falls back below the inhibitory threshold (blue block in Fig.~\ref{fig:survival_map}A(2)).
Their ratio defines the \emph{recovery ratio} $R$, and clearance requires
\begin{equation}
0 < t_{\mathrm{i}}/t_{\mathrm{r}} = R < (N-1),
\label{eq:clearance_cond}
\end{equation}
where $N$ is the number of modules in the ring and $t_{\mathrm{i}}$ the total time this module is above the inhibitory threshold per cycle.
Equation~\ref{eq:clearance_cond} applies only when all three pathways are present; when inhibition is absent, $t_{\mathrm{i}}$ is undefined and saturation occurs regardless of timing parameters, as illustrated in Fig.~\ref{fig:survival_map}B(1).
These two are distinct conditions calculated from the same module timing, and the oscillatory region is their intersection.
Because the clearance bound scales as $N-1$, the oscillatory region widens with network size.

{Evaluation of both conditions across parameter space produces a two-parameter bifurcation diagram, in which a bounded oscillatory region is surrounded by extinction (steady-state uninflated) and saturation (steady-state inflated) zones (Fig.~\ref{fig:survival_map}).
}
The full parameter space spans five dimensions ($th_e$, $th_i$, $\tau_{\mathrm{on}}$, $\tau_{\mathrm{off}}$, and $N$), so we present two representative cross-sections.

The first cross-section fixes the thermal time constants and network size, and varies the threshold separation to analyze drift under mechanical load such as adjusted switch threshold or change in the total vapor pressure (Fig.~\ref{fig:survival_map}A).
Stable oscillation occupies a bounded wedge.
The threshold separation $th_e - th_i$, referred to here as the guard window, sets the timing margin between excitation and subsequent inhibition.
When the guard window is too narrow, modules are inhibited almost immediately after exciting their successors, violating the clearance condition and driving the system into saturation. When the guard window is too wide, the recovery time grows beyond what the ring's transit time can provide, violating clearance conditions.

The second cross-section fixes the thresholds and network size, and varies the thermal time constants in environmental parameters such as cooling and heating as well as the input power (Fig.~\ref{fig:survival_map}B).
When cooling is fast relative to heating (small $\tau_{\mathrm{off}}/\tau_{\mathrm{on}}$), modules recover quickly but lose their excitatory output before successors can latch, leading to extinction (Fig.~\ref{fig:survival_map}B(3)), located in the lower right corner.
When cooling is slow relative to heating, modules remain active too long, and the returning wave collides with modules that have not yet recovered.
This slow cooling produced saturation as the red zone in the top left region and a sample trace can be seen in Fig.~\ref{fig:survival_map}B(1).
A strip-shaped oscillatory region separates these two failure modes.
In the following sections we describe a series of perturbation-related experiments, where
each physical perturbation tested corresponds to a specific trajectory on this map.

\subsection{\name as an LIF-inspired excitable module}
\label{sec:design_regenerative_dynamics}

\begin{figure}[h!]
 \centering
 \includegraphics[width=1\linewidth]{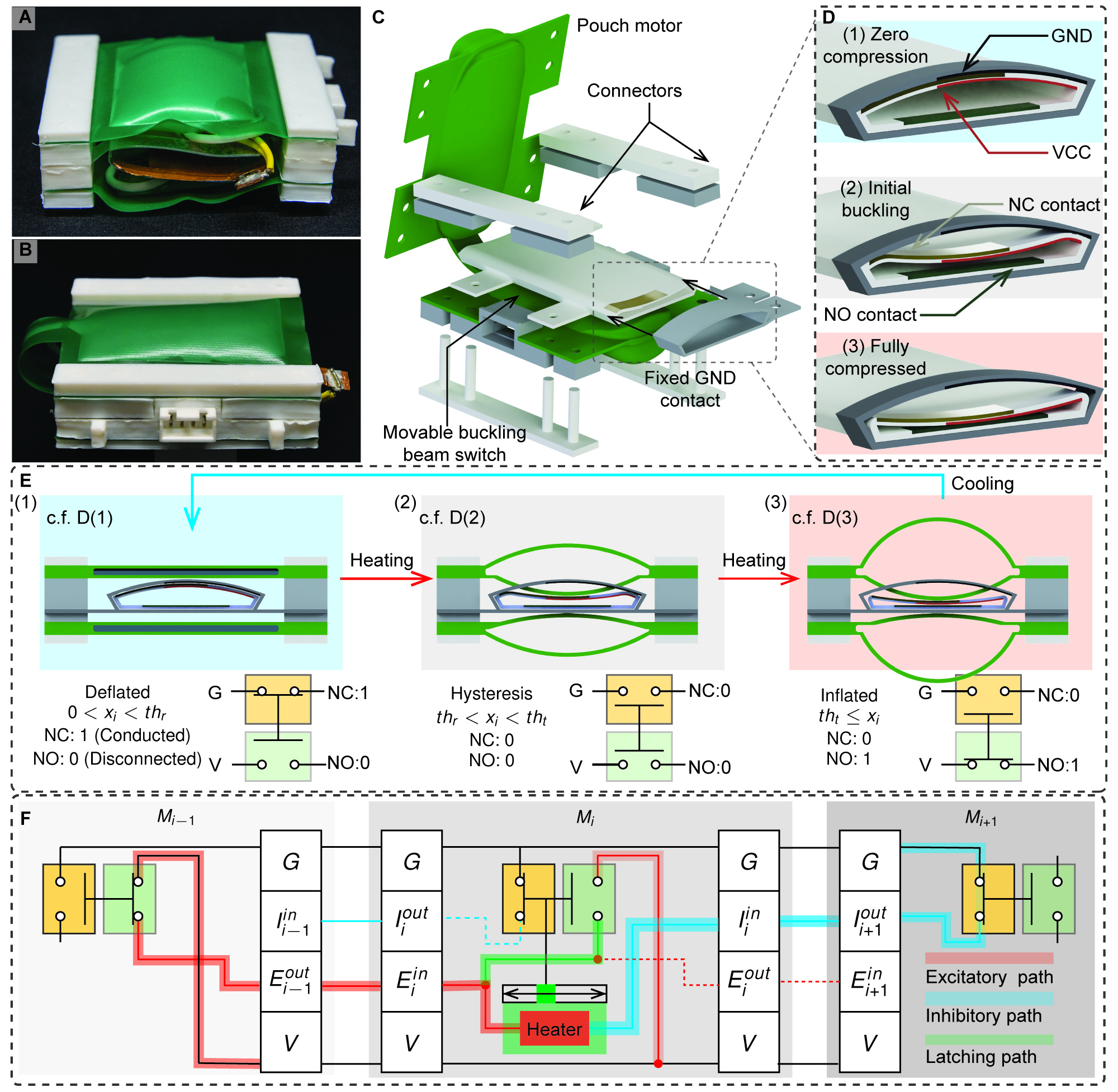}
 \caption{
Detailed design of a single \name module.
(A), (B) Physical prototype.
(C) Exploded rendering of a single module with the pouch motor unfolded to highlight internal design.
(D) Mechanical compression sequence of the two-part buckling switch.
(E) Thermal activation sequence of the actuator with cross-section, along with the switch configuration at the bottom.
(F) Equivalent logic representation showing excitatory ($E_i$), inhibitory ($I_i$), and self-latch ($S_i$) pathways, as well as the ground ($G$) and power supply ($V$). $\m{i}$ stands for the $i$-th module in the system.
 }
 \label{Fig:single_module}
\end{figure}

\S~\ref{sec:chain_propagation} and \S~\ref{sec:survival_map} together specify a candidate architecture for soft excitable networks, where each module must implement three signaling pathways and operate inside the time range bounded by the regeneration and clearance conditions.
We now describe a physical realization of such a module that satisfies both requirements using a single integrated mechanical switch (Fig.~\ref{Fig:single_module}).

The module is built around a dual-chamber thermoplastic polyurethane (TPU) pouch pre-filled with a low boiling-point fluid (LBF)~\cite{niiyama_pouch_2015,mirvakili_actuation_2020,liu_ethanol_2021, gockowski_improving_2025,han_untethered_2019}, wrapped around a central two-part buckling-beam switch. 
When power is applied, the heaters increase the temperature of LBF, inducing a liquid-vapor phase transition that generates internal pressure and pouch inflation (Fig.~\ref{Fig:single_module}D).
The internal pressure increase is continuously offset by passive thermal dissipation to the environment.
Strain-limiting connectors along the longitudinal edges direct the isotropic inflation into targeted vertical bulging coupled with linear contraction.
This deformation applies progressive compressive force to the internal switch, coupling the thermodynamic state of the LBF to electro-mechanical switching.

The switch provides one normally closed (NC) contact and one normally open (NO) contact, defining three electrical states governed by mechanical deformation (Fig.~\ref{Fig:single_module}D).
In the idle state, only the NC contact is engaged with the ground (GND) while the NO contact is disconnected from voltage at the common collector (VCC).
With initial inflation, the TPU pouch applies compression to the switch body, disengaging the buckling beam from the outer ring and thus the NC contact.
When a module's state of activation exceeds $th_i$, it enforces shutdown of the previous module by disconnecting that module's heater from the power rail through the inhibitory path.
We define this as the inhibitory threshold, $th_i$. Before the compression is sufficient to conduct the NO contact, which we refer to as the excitatory threshold $th_e$, the switch enters an intermediate regime where both NC and NO contacts are disengaged.
The separation between $th_i$ and $th_e$ creates a guard window, providing temporal overlap between signal reception and transmission.
With further inflation, once the excitatory threshold $th_e$ is reached, the conducted NO contact activates the successor module through the excitatory path.

The self-latching pathway is implemented through local routing on the same NO contact. The buckling switch itself returns to its rest position once the pressure is released, providing no mechanical bistability, so the logical memory required by the architecture in \S\ref{sec:chain_propagation} must come from electrical routing. We connect the module's own NO contact in parallel with the upstream module's NO contact, both feeding the VCC of the same heater (Fig.~\ref{Fig:single_module}F). Before the module crosses $th_e$, its own NO has not yet conducted, and the heater is powered only by upstream excitation. Once the module crosses $th_e$, its NO closes and supplies power to the heater in parallel, maintaining activation independently of the upstream module until being shut down by its successor.

\begin{figure}[h!]
 \centering
 \includegraphics[width=0.99\linewidth]{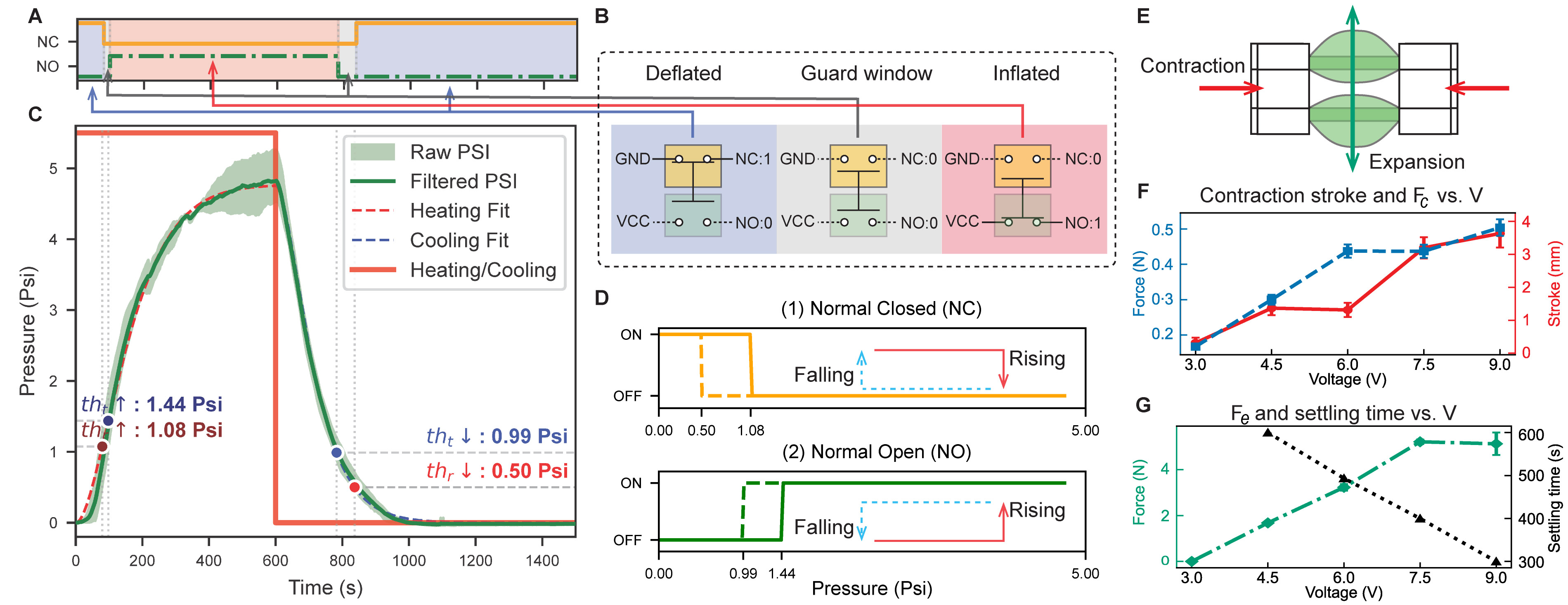}
 \caption{
 \textbf{Single-module pressure dynamics and mechanical output.}
 (A) Conduction state of the switch over a complete heating-cooling cycle;
 (B) Digital configuration of the switch under load. Shaded background colors indicate the position in status bar in (A), as well as the activation state in Fig. 2(E);
 (C) Pouch pressure and measured mechanical threshold;
 (D) Hysteresis of the switch;
 (E) Contraction and Expansion of the pouch and direction of the force; 
 (F) Lateral force and stroke at steady state versus voltage;
 (G) Vertical force and required settling time versus voltage.
 }
 \label{fig:single_char}
\end{figure}

The circuitry of a connected \name places three contacts around the heater in $\m{i}$, including the NC from $\m{i+1}$ as inhibitory, NO from $\m{i-1}$ as excitatory, and NO from itself as latching (Fig.~\ref{Fig:single_module}F). The NC contact at $th_i$ controls the GND end of the upstream module's heater, providing the inhibitory pathway. The NO contact at $th_e$ feeds the VCC end of both the upstream module's path and the module's own heater, providing the excitatory and self-latching pathways, respectively. This dual-threshold switch with local path satisfies the three pathway requirements identified in \S\ref{sec:chain_propagation}.

\subsection{Characterization of a single Pneu-ron.}
\label{sec:baseline_char}

To establish baseline parameters for network analysis, we measured pressure dynamics during a full activation-deactivation cycle and observed that the pressure exhibited a second-order under-damped response (Fig.~\ref{fig:single_char}A--C).
Since the switch was made of compliant material, we observed hysteresis between heating and cooling cycles, yielding ($th_i^{\uparrow}$, $th_i^{\downarrow}$) for the inhibitory threshold ($th_i$) and ($th_e^{\uparrow}$, $th_e^{\downarrow}$) for the excitatory threshold ($th_e$), during rising ($^{\uparrow}$) and falling($^{\downarrow}$), respectively (Fig.~\ref{fig:single_char}D).
These parameters at room temperature establish the baseline performance of the \name as a stable excitable element, forming the quantitative foundation for analyzing network behavior in subsequent sections.

During activation, the pouch actuator produced two motions: linear contraction along the longitudinal axis (red arrows in Fig.~\ref{fig:single_char}E) and lateral expansion perpendicular to it (green arrows).
At steady state, contraction force ($F_c$) and stroke increased monotonically with input voltage (Fig.~\ref{fig:single_char}F).
At 9\,V, the module achieved a maximum contraction force of approximately 0.5\,N with a stroke of 4\,mm.
The expansion force ($F_e$), which drives the internal switch mechanism, showed a similar voltage dependence (Fig.~\ref{fig:single_char}G), reaching approximately 4\,N at 9\,V. 
Settling time decreased with increasing voltage, from approximately 600\,s at 4.5\,V to 300\,s at 9\,V.

\begin{figure}[h!]
 \centering
 \includegraphics[width=1\linewidth]{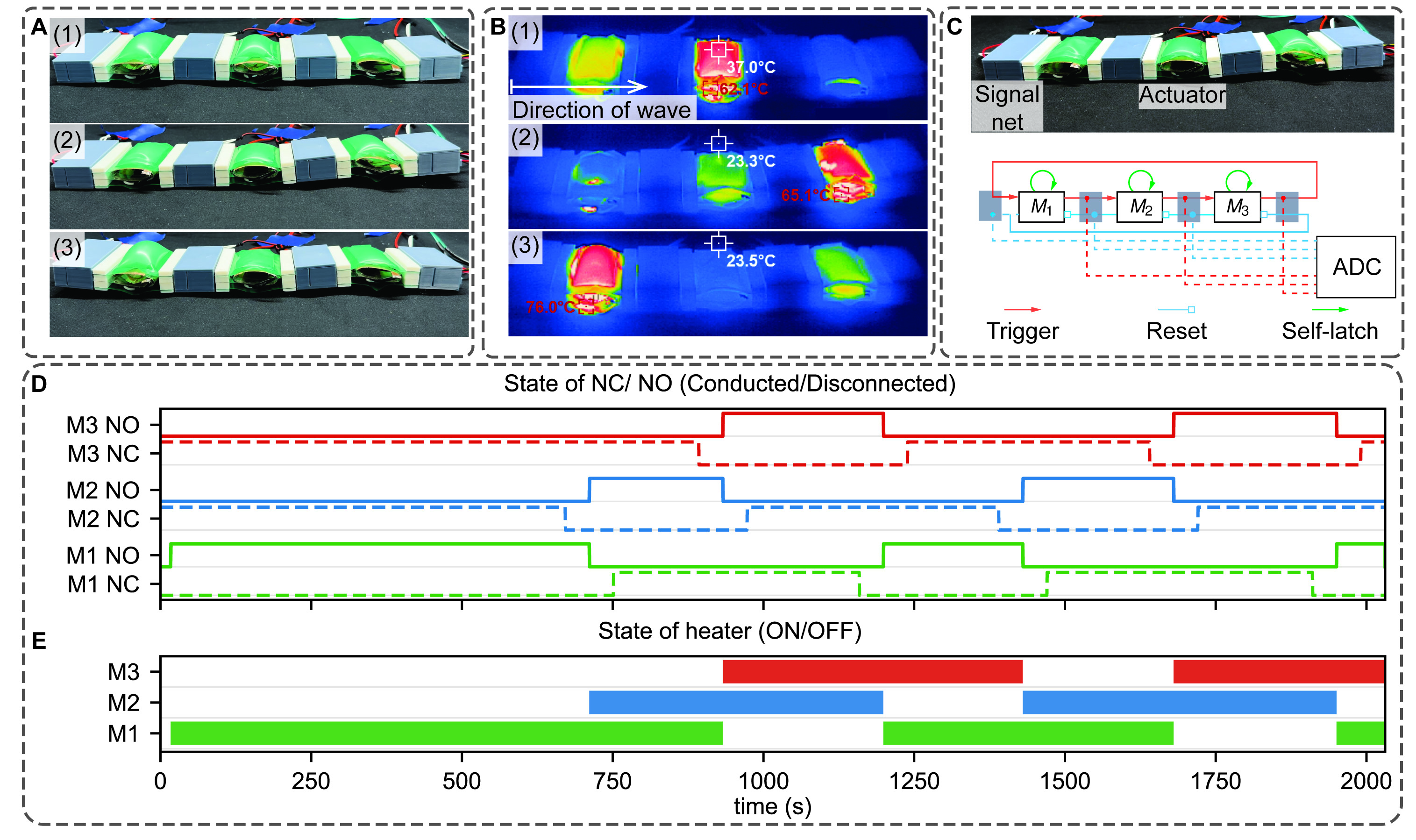}
 \caption{
 \textbf{Oscillation of a CPG ring and its experiment data.}
 (A) Experiments of 3-segment wave propagation;
 (B) Thermal image showing three states;
 (C) Data collection setup illustration.
 (D) State of the NC and NO switch during a typical cycle of a 3-segment prototype;
 (E) Calculated conduction state of the heaters during a typical cycle.
 }
 \label{fig:state_exp}
\end{figure}

\subsection{Three-segment CPG rings sustain oscillation}
\label{sec:state_exp}

With the single-module parameters established, we assembled \name modules into rings to test the bifurcation diagram predictions experimentally.
A 3-segment ring produced sustained traveling-wave oscillation at room temperature, with switch outputs and thermal images confirming the ``2-hot'' handoff pattern predicted by the network analysis (Fig.~\ref{fig:state_exp}A--B).
At any moment, two heaters are simultaneously conducting, with one module above the $th_e$ and its successor rising toward the excitatory limit (Fig.~\ref{fig:state_exp}D--E).

\begin{figure}[h!]
 \centering
 \includegraphics[width=1\linewidth]{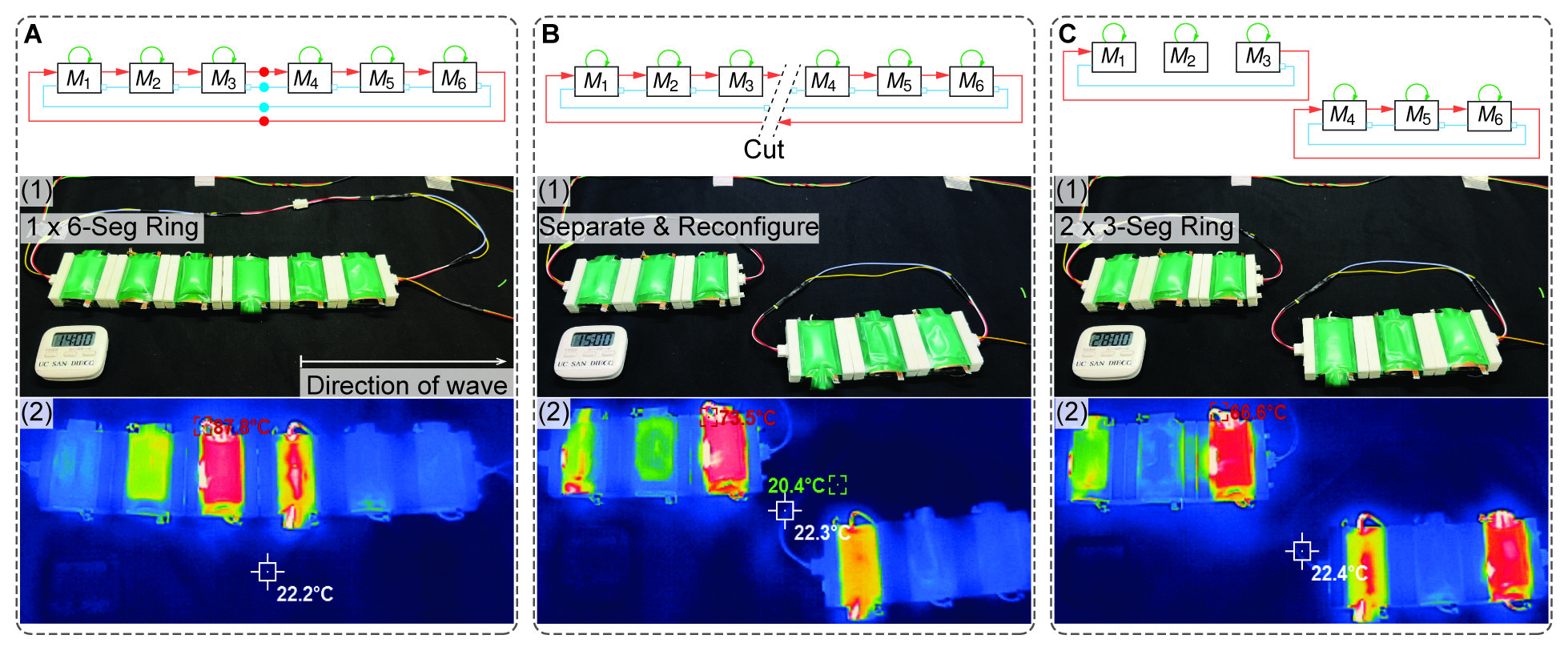}
 \caption{
 \textbf{Reconfiguration experiments.}
 (A) 6-segment topology; (B) Separation of  6-segment in the middle; (C) Reconnection generates two rings; Inside each panel, all sub-panels of (1)s are photos while (2)s are thermal images.
 }
 \label{fig:reconfig}
\end{figure}

\subsection{Demonstration of network reconfiguration}
\label{sec:reconfig_exp}

To test whether oscillation persists across topology changes, we performed a reconfiguration experiment on a 6-segment ring.
The ring was first activated at 9.0~V for 14~minutes to allow one complete cycle of oscillation, with the wave reaching $\m{3}$ a second time (Fig.~\ref{fig:reconfig}A).
We then physically separated the ring at the $\m{3}/\m{4}$ and $\m{6}/\m{1}$ junctions during active wave propagation, creating two independent 3-segment sub-rings: $\m{1}$--$\m{3}$ and $\m{4}$--$\m{6}$ (Fig.~\ref{fig:reconfig}B).
Because $\m{4}$ had not yet received sufficient excitation from $\m{3}$ at the moment of disconnection, it failed to enter the self-latching state and lost its excitation, as visible in the thermal image immediately after the cut (Fig.~\ref{fig:reconfig}B). Meanwhile, $\m{3}$ completed its cycle and immediately re-excited $\m{1}$ in the first sub-ring. A brief mechanical compression was applied to $\m{4}$ to manually restart the second sub-ring.

Both 3-segment rings resumed oscillation within one cycle, producing an independent traveling wave at a longer per-module time than the original 6-segment ring (Fig.~\ref{fig:reconfig}C).
{This is because each handoff in a 3-segment ring needs a longer time for the previous wave to be cleared 
(See \S5 of the Supporting Information).}
No reprogramming or parameter adjustment was required beyond the manual excitation of $\m{4}$. The thermal images confirm independent wave activity in both sub-rings via the ``2-hot'' handoff pattern (Fig.~\ref{fig:reconfig}C(2)), with hot spots of differing intensity indicating sequence of the phases.
The two sub-rings continued to operate autonomously for an additional 13~minutes until the $\m{4}$--$\m{6}$ ring's wave returned to the front, confirming oscillation in both sub-rings.
This single experiment also places two network sizes (N=6 and N=3) inside the predicted oscillatory region simultaneously.
We report this as a proof-of-principle demonstration for ideal, symmetric cutting and propose arbitrary cut locations as future work.

\subsection{Response to mechanical perturbations}

\begin{figure}[tb]
 \centering
 \includegraphics[width=1\linewidth]{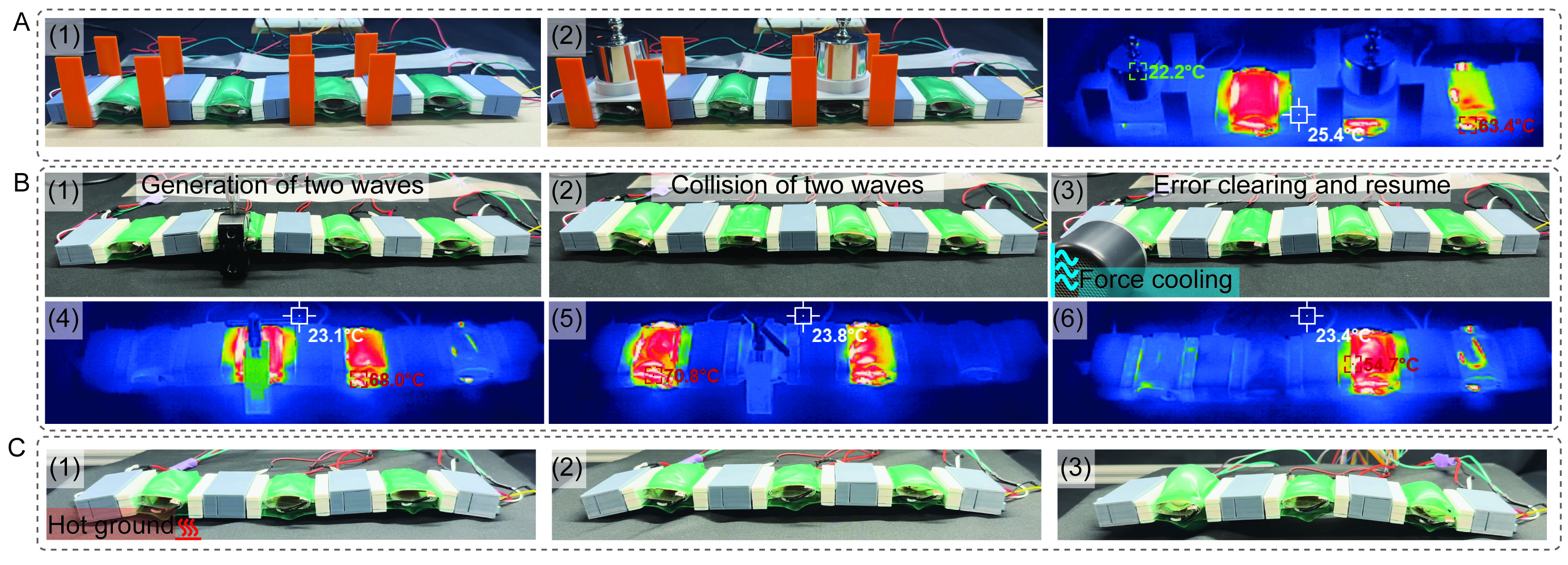}
 \caption{\textbf{Emerging behavior under shifted environmental parameters.}
 (A) Weight loading experiment; (1) is the unloaded condition while two 500~g weights are applied in (2). Thermal images in (3) indicate continuing wave propagation.
 (B) Conduction block sequence; (1-3) overview of the experiments, (4-6) corresponding thermal images.
 (C) Hot-plate experiments. (1-3) are the time sequence.
 }
 \label{fig:neuron_P}
\end{figure}

The bifurcation diagram predicts that a mechanical perturbation along each axis produces the specific failure mode indicated by the nearest boundary.
We tested these predictions by applying mechanical loads, both affecting the excitatory and inhibitory thresholds to running CPG rings, as seen in Fig.~\ref{fig:neuron_P}(A).
Mechanical loading altered the effective switching thresholds of the \name by changing the compressive force required to actuate the buckling beam.
On the bifurcation diagram, this corresponded to motion along the threshold axis of Fig.~\ref{fig:survival_map}A.

With 500~g applied to each of $\m{1}$ and $\m{3}$ (20.8$\times$ body weight of 24~g) the 4-segment ring continued to oscillate, with reduced activation time on the loaded modules, seen in the photos Fig.~\ref{fig:neuron_P}A(1--2). The thermal image in  Fig.~\ref{fig:neuron_P}A(3) confirms this oscillation.
The rising time, defined as the latency for $\m{i}$ to issue its excitatory signal to $\m{i+1}$, decreased by {${\sim}40\%$} on the loaded modules (e.g.\ $\m{1}$: ${\sim}{25}$ to ${\sim}{15}$~s).
This change in the timing behavior is because the added compression reduces the switching threshold and shortens the guard window $\Delta\mathit{th}$, a downward, sensitization-like shift of the operating point that remains within the oscillatory region of the bifurcation diagram (Fig.~\ref{fig:survival_map}A).
The period, by contrast, increased by {${\sim}34\%$} ({632} to {848}~s).
Each loaded module fired sooner, but the weight resists the deflation that would release it, so the module stayed active far longer.
Since the ring advances only when the active module releases, this slow release sets the pace of the whole cycle.
Loading therefore leaves a module quick to fire but slow to release, and the rhythm slows without breaking.

Localized clamping to a single module in a ring produced a conduction block (Fig.~\ref{fig:neuron_P}B).
The clamp forced the NO contact, the excitatory pass in $\m{2}$, meaning that $\m{2}$ was permanently sending excitatory signals to $\m{3}$ while never inhibiting $\m{1}$, which disrupted the normal wave sequence.
The traveling wave propagated from $\m{2}$ to $\m{4}$ and fed back to $\m{1}$ as usual, but because $\m{1}$ received no inhibitory signal from the clamped $\m{2}$, it remained self-latched in the ON state.
With $\m{2}$ mechanically locked and $\m{1}$ permanently active, $\m{3}$ was re-activated by $\m{2}$'s constant excitatory output. Thus two waves in a 4-segment ring were generated and the system then settled into a state where only $\m{1}$ and $\m{3}$ remained permanently active while $\m{2}$ and $\m{4}$ were permanently off, as seen in Fig.~\ref{fig:neuron_P}B(2,4).
In effect, two excitatory signals co-existed in the ring where the two waves collided and locked.
The system remained stuck in this state even after the external load was removed, because the self-latch on $\m{1}$ and $\m{3}$ maintained their activation without any mechanism to break the deadlock.
For $N=4$ at $R=2$, the clearance condition allows only $N/(R+1)\approx1.3$ waves numerically, which is a single traveling wave.
When a second wave appeared, these two waves collided and caused deadlocks.

This collision-induced conduction block in a limited size ring was reversible.
Forced air cooling of $\m{1}$ dropped its temperature below the inhibitory threshold, breaking its latching.
With $\m{1}$ reset, the restriction on $\m{4}$ was lifted, and only one wave survived. The system then resumed its normal single-wave traveling pattern (Fig.~\ref{fig:neuron_P}B(3,6)). A demonstration of this process can be seen in Movie S4 of the Supporting Information.
These results suggest that the \name network exhibits failure modes that emerge from the interaction between topology and threshold mechanics, not from material failure, and that these failures are reversible through thermal reset that governs normal operation.

\subsection{Thermal sensitivity and perturbations}

Similar to the experiments of mechanical perturbation, thermal perturbation shifts the operating point along the thermal axis of Fig.~\ref{fig:survival_map}B.
Reducing ambient temperature increases $\tau_\mathrm{on}$ and decreases $\tau_\mathrm{off}$, driving the system towards extinction.
On the opposite side, increasing $\tau_\mathrm{off}$ while reducing $\tau_\mathrm{on}$ drives toward saturation. Decreasing the input voltage provides a similar effect, where it increases $\tau_\mathrm{on}$ while leaving $\tau_\mathrm{off}$ nearly unchanged.
These three perturbations together sample three distinct trajectories through Fig.~\ref{fig:survival_map}B's thermal region.
At $-20\,^{\circ}\mathrm{C}$, the traveling wave extinguished within seconds as the rapid heat loss to the cold environment prevented any module from reaching the excitatory threshold (Fig.~\ref{fig:survival_map}B).
This corresponds to a diagonal shift from the oscillatory region toward the lower-right extinction zone in Fig.~\ref{fig:survival_map}B from the baseline green circle to the blue cross, driven by the simultaneous increase in $\tau_{\mathrm{on}}$ and decrease in $\tau_{\mathrm{off}}$.

Reducing supply voltage from $9.0\,\text{V}$ to $7.5\,\text{V}$ produced extinction through a different trajectory. After the voltage reduction, the system completed approximately one additional cycle, but the signal amplitude at each module decayed progressively, and the wave eventually extinguished.
Because voltage reduction slows internal heating without significantly affecting passive cooling, this perturbation increases $\tau_{\mathrm{on}}$ while holding $\tau_{\mathrm{off}}$ nearly constant.
This is shown as a horizontal rightward shift from the oscillatory region into the extinction zone in Fig.~\ref{fig:survival_map}B from baseline to purple cross.

On a $60\,^{\circ}\mathrm{C}$ hotplate that exceeds the fluid's $34\,^{\circ}\mathrm{C}$ boiling point, the ring saturated, as seen in Fig.~\ref{fig:neuron_P}C(1). 
Because ambient heat drives the fluid beyond its phase-change limit, the operating point stays outside the model's validity, and we treat it as a qualitative saturation case.
Modules $\m{1}$ and $\m{3}$ inflated spontaneously under the ambient heat. Because the internal switch structure acts as a strain limiter, each inflated module curled into a C-shaped configuration rather than expanding uniformly.
The combined curvature of $\m{1}$ and $\m{3}$ formed an arch that lifted the $\m{2}$ off the hotplate surface, as seen in Fig.~\ref{fig:neuron_P}C(2).
Elevated above the heat source, $\m{2}$ did not receive sufficient thermal input to vaporize and remained in an idle state.
Although $\m{1}$ was sending an excitatory signal to $\m{2}$, $\m{3}$ was simultaneously sending an inhibitory reset, keeping $\m{2}$ nonconducting.
This deviation from normal signaling behavior resulted in a stable, static arch, where $\m{1}$ and $\m{3}$ remained permanently inflated and $\m{2}$ was suspended between them.
More interestingly, since $\m{2}$ was in the idle state, its NC contact on the switch allowed $\m{1}$ to be activated, while the NO contact of $\m{3}$ was activated by heat.
This configuration of the switch caused an electrical conduction on $\m{1}$, making $\m{1}$ get both ambient and electric heating. As a result, the inflation amplitude of $\m{1}$ was significantly larger than the other modules, Fig. ~\ref{fig:neuron_P}C(3).
A demonstration of this process can be seen in Movie S5 of the Supporting Information.
This state corresponds to low $\tau_\mathrm{on}$ and high $\tau_\mathrm{off}$, placing the operating point in the upper-left saturation zone in Fig.~\ref{fig:survival_map}B from baseline green dot to the red cross on the upper left corner.
The boiling point of the working fluid (34°C) means any ambient temperature above this value spontaneously activates the modules.

\begin{figure}[h!]
 \centering
 \includegraphics[width=1\linewidth]{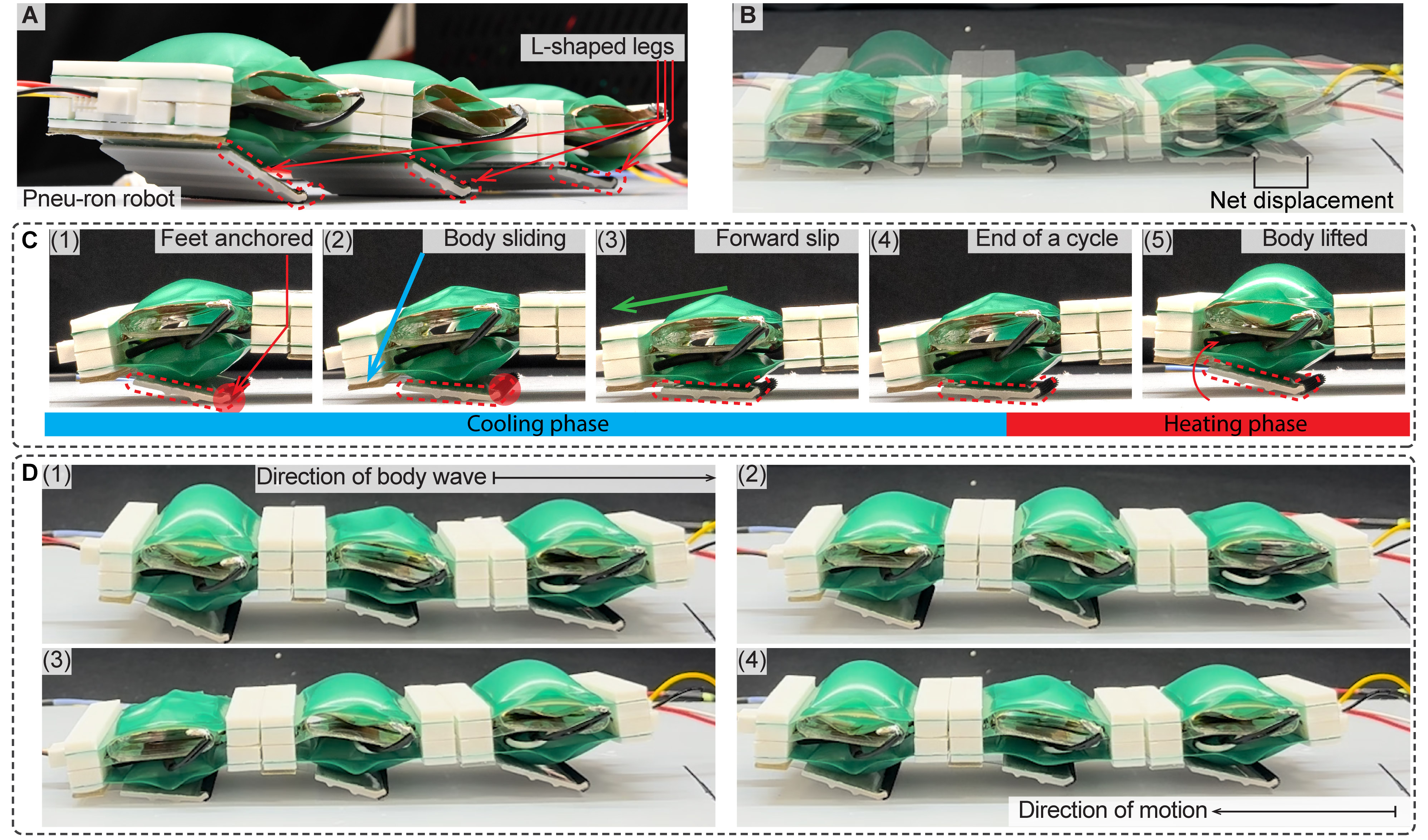}
 \caption{
 \textbf{Peristaltic locomotion driven by traveling-wave CPG.}
 (A) Assembled \name ring with leg for locomotion. Red bounding boxes indicate the L-shaped legs attached to the bottom of each module.
 (B) Overlay image of the robot at the beginning and in the end to highlight net displacement.
 (C) Crawling sequence of the robot. The claw anchors in frame 5 in the heating phase, and generates forward slip for movement during the cooling cycle. 
 (D) Full robot crawling sequence with claw in a full cooling and heating cycle.
 }
 \label{fig:locomotion_crawl}
\end{figure}

\subsection{Locomotion by mechanical coupling of the traveling wave}
\label{sec:locomotion}

Coupling the traveling wave to legs with state-dependent friction converts the network activation sequence into peristaltic crawling, without any additional timing signal.
In a closed \name ring, the body wave is a network-level sequence in which one module inflates while its predecessor deflates.
We tested this coupling with a minimal 3-segment ring assembled into a worm-inspired robot on a flat acrylic surface (Fig.~\ref{fig:locomotion_crawl}A).

Each module carried an L-shaped leg, with a short frictional claw at the distal end and a longer lever arm (Fig.~\ref{fig:locomotion_crawl}A).
When module $\m{i}$ inflates at the end of the heating phase (Fig.~\ref{fig:locomotion_crawl}C(5)), the distal end of the leg pivots about the ground.
The frictional claw then engages the ground and anchors the segment through increased friction (Fig.~\ref{fig:locomotion_crawl}C(1-2)).
During deflation, only the long arm contacts the ground, which provides low-friction sliding support (Fig.~\ref{fig:locomotion_crawl}C(3-4)).
This asymmetry between the inflated and deflated states converts vertical inflation into directional ground engagement.

Sequential activation converts the per-module asymmetry into net displacement of the whole ring. 
During inflation, the next inflating module, $\m{i+1}$, pushes its released neighbor, $\m{i}$, in the direction opposite to wave propagation, and the ring advances by one increment per activation cycle (Fig.~\ref{fig:locomotion_crawl}D).
The resulting gait follows the reverse peristalsis of earthworms, in which a backward-traveling wave produces forward displacement~\cite{quillin_kinematic_1999}.
No gait generator or actuator timing was required, because the same inflation event both actuated a segment and triggered its successor.

Over six cycles, the 180-mm-long robot advanced 25~mm, equivalent to 0.0231 body lengths (BL) per cycle (Fig.~\ref{fig:locomotion_crawl}B).
The cycle period of approximately 15~minutes reflects the thermal dynamics of liquid-vapor phase change in the chosen working fluid, combined with the low-power heaters selected for safe operation.
The displacement per cycle falls within the range of 0.02 to 0.06 BL per cycle observed in biological earthworms~\cite{quillin_kinematic_1999}, where the same peristaltic strategy operates at faster timescales.
Speed here is a property of the implementation rather than the architecture.
Using alternative working fluids or higher-power heating modalities would shift the absolute timescale without changing the underlying excitable network behavior.

\section{Discussion}
\label{sec:dis}
In this work, we developed soft robotic pneumatic neuron modules (Pneu-rons) that use low-boiling-point fluid (LBF) inflation for pressure generation, actuation and internal signaling. 
By incorporating a soft mechanical switch within each module, the inflation actuation can trigger electrical signals to adjacent modules. 
A minimal dynamics model of inflation inspired by the integrate-and-fire model of neural networks enabled us to predict how network topology and module properties (inflation time, triggering threshold, etc.) can yield robust steady-state oscillations. 
Experimental tests of \name networks demonstrated modularity, robustness to damage, environmental adaptation of oscillator properties, and worm-like locomotion using frictional feet.

One of the many advantages of soft robots is their resilience to damage~\cite{tolley_resilient_2014,shepherd_multigait_2011}.
Since soft robots are inherently made from compliant materials, they can sustain large stresses and strains and still remain operational. 
However, soft robots that use pneumatic actuation can be extremely sensitive to damage if the pressurized air channels are punctured or cut.
In this work we demonstrated that a worm-like robot, when severed in half, can still maintain oscillatory function in both of the halves, in effect creating two new robots.
This robustness to damage is a testament to the modular design paradigm of Pneu-rons.
Since each module is self-contained and inflates through the low-boiling-point fluid, the system does not need a pump and so can be resilient to damage that may otherwise cause a robot to fail. 
An exciting area of future research will be to better understand how network topology and mechanical properties of the Pneu-rons impact robustness and resilience. 

Inflation oscillations in \name networks generated traveling waves of actuation along the robot body. 
The speed of these traveling waves is sensitive to the mechanical and thermal properties of the module, and additionally the external environment acts as a \textit{participant} in the dynamics rather than a \textit{disturbance}.
Thus, soft robots created from networks of pneumatic neurons are able to adapt actuation dynamics dependent upon the external load, ambient temperature and environment (submerged versus in air for example), and overall energy input. 
This sensitivity to the environment can enable new opportunities for designing adaptive soft robots that change their function when experiencing external loads or variations in external temperature.
A feature of each \name is that this adaptability can be achieved without a sensor, force transducer, or digital computation, which makes these networks a physical instance of distributed, morphological control rather than a learned or computed policy~\cite{hauser_towards_2011,tanaka_recent_2019,sitti_physical_2021}.

While low-boiling-point fluids enable modular pneumatic actuation without a pump, there are several challenges and opportunities that remain. 
For example, the working fluid and heater were contained in a pouch of TPU material which was selected for its sealability and airtightness.
However, TPU is sensitive to heating and this limited the maximum heater temperature,  which in turn limited the maximum speed of inflation. 
In our earlier designs, we did explore using silicone for the pouches. However, we found that the LBF would easily leak through diffusion and swelling, and needed to be replenished over time~\cite{miriyev_soft_2017}.
Thus, the next advancements in Pneu-rons will be enabled by material selection and design innovations that balance thermal input (for actuation speed), inflation volume (for larger actuation displacement), and optimal materials (that enable reliable fabrication and robust operation).

The self-contained architecture of \name could support soft devices for underground inspection and locomotion, as well as underwater and space exploration.
It could also support other applications where pneumatically driven soft robots are preferred but an external gas supply is challenging to provide.
Beyond driving their own bodies, \name networks could serve as physical pacemakers for other robotic units or thermally driven end effectors.
The hotplate experiments suggest that excessive temperatures could be encoded as persistent electrical signals given the network topologies.
Such functions would use the same module architecture to connect environmental responses with signaling and actuation.

The centralization of control is a common design paradigm in robotics where a single electrical controller sends actuation commands across the body. 
However, decentralized controllers have some advantages such as modularity and parallelization. 
Running on a single DC supply with no controller, clock, or feedback sensor, the \name ring produced sequential peristaltic motion.
By embedding actuation, logic, and timing directly into one compliant body, \name modules blur the boundaries between actuator and controller, and between muscle and neuron.
The excitable-network template and its bifurcation diagram give a candidate substrate-agnostic framework for designing, analyzing, and comparing self-oscillating systems.

\section{Materials and Methods}

\subsection{Mechanical design of the pouch actuator}
\label{sec:M_design}

Each \name comprises two sub-assemblies: a fluid-filled pouch that converts heat into pressure and a buckling switch that converts that pressure into electrical contact changes.
The pouch is a dual-chamber thermoplastic polyurethane (TPU) envelope pre-filled with a low boiling-point fluid (LBF, Opteon SF33).
Thirty centimeters of silicone-coated heaters, with a resistance of 125$\Omega$ per meter, embedded on the inner wall serves as the energy input.
The pouch is wrapped around a central, two-part internal buckling-beam switch structure (Fig.~\ref{Fig:single_module}A-C).
The switch is folded from a laser-cut plastic sheet into a pentagon-like three-dimensional structure, with copper tape applied as conductive traces and adhesive layers.
Both ends of each NC and NO contact sets are soldered to the JST connectors on the side, and the JST electrical terminals are then embedded in a custom plastic mechanical connector.
The buckling geometry causes the NC contact disengage before NO can be contacted and sets the thresholds $th_i$ and $th_e$. Details regarding manufacturing method and material selection are provided in \S1 of Supporting Information.

\subsection{Characterization methods}
\label{sec:M_characterization}
We characterized a single \name in three ways: pressure and switching behavior over a full cycle, the time constants fitted from that response, and mechanical output against supply voltage.
We first recorded internal pouch pressure and digital switch outputs during complete heating-cooling cycles at room temperature (Fig.~\ref{fig:single_char}A,B).
Pressure was chosen because it is a direct observable of the coupled thermo-fluidic-mechanical state.
A custom \name with a thin channel connected to a pressure sensor (Adafruit MPRLS Ported Pressure Sensor) was manufactured. Since only one module is used in this experiment, we directly connected the ends of the heaters to a DC power supply for direct activation. 
We then measured the pressure thresholds at which the NC and NO contacts change conduction state. 
We fitted the recorded pressure to a second-order RC-circuit model with separate heating and cooling branches. The fitted form, its parameters, and its simplification for network simulation are given in \S2 of the Supporting Information.

We also measured the mechanical output of a single \name under varying voltage inputs at the steady state (Fig.~\ref{fig:single_char}E--G) using a load cell (ATI Delta IP68). The actuator was connected to the load cell using a tendon with springs and we synced the heating power with the load cell to obtain force reading during data collection.
Additionally, to track the contraction of the actuator in the lateral direction, a reflective marker was attached to the plastic connector on the side. A motion capturing system (Motive OptiTrack) was used to measure the change in location in the world frame.

\subsection{Network simulation}
\label{sec:M_simulation}

We model each module as a first-order RC circuit with two signaling thresholds, the inhibitory threshold, $th_i$ and the excitatory threshold $th_e$.
Although the physical system exhibits second-order dynamics with hysteretic switching, we simplified the simulation to a first-order RC model with fixed threshold values.
The qualitative failure modes are governed by the ratio of thermal time constants rather than the oscillatory transient, so this approximation captures the essential dynamics.
During simulation, an external stimulus briefly activates $\m{1}$, and we observe how the signal propagates through the chain under four network configurations (Fig.~\ref{fig:signal_speed}C--F).

The bifurcation diagram is evaluated using the same model.
Its parameter space spans the two thresholds ($th_e$, $th_i$), the two thermal time constants ($\tau_{\mathrm{on}}$, $\tau_{\mathrm{off}}$), and the network size $N$, so we present two representative two-dimensional cross-sections in Fig.~\ref{fig:survival_map}.
The first fixes the thermal constants and network size ($\tau_{\mathrm{on}}=215$, $\tau_{\mathrm{off}}=118$, $N=4$) and varies the threshold separation.
The second fixes the thresholds and network size ($th_i=0.21$, $th_e=0.28$, $N=3$) and analyzes the two thermal time constants.

\subsection{Experimental protocol of perturbation test}
\label{sec:M_perturbation}
We investigated how mechanical loading affects CPG dynamics by applying controlled loads to a running 4-segment ring while recording the switch output of each module. 
Two protocols were used, targeting different regions of the bifurcation diagram in Fig.~\ref{fig:survival_map} threshold axis.
In the first protocol, two 500~g weights were placed on modules $\m{1}$ and $\m{3}$ during sustained oscillation.
In the second protocol, a mechanical clamp was applied to $\m{2}$, forcing the NO contact in the conducted state. Recovery was initiated by directing forced air from a fan onto $\m{1}$.

In the thermal perturbation experiments, a normally operating 3-segment ring at room temperature was placed directly into a freezer at $-20\,^\circ$C.
The supply voltage was reduced from 9.0~V to 7.5~V during sustained oscillation for the voltage reduction experiments.
During the demos of hot saturation, a 3-segment ring was placed on a hotplate set to $60\,^\circ$C, while connected to power.
In this test in particular, no manual trigger was applied and we used heating from the ambient to actuated the robot.

\subsection{Experimental protocol of reconfiguration test}
\label{sec:M_reconfig}
Unlike the perturbations above, which displace the operating point without changing the network, this experiment reconfigures the topology itself.
When reconfiguration experiments were conducted, a 6-segment ring was first operated at 9.0~V for 14~minutes to allow one complete cycle of oscillation.
Once the oscillation signal arrived at $\m{3}$, the ring was physically separated at the $\m{3}$/$\m{4}$ and $\m{6}$/$\m{1}$ junctions during active wave propagation. 
Module $\m{4}$ was then manually restarted by brief mechanical compression. Both sub-rings were monitored for an additional 13~minutes.

\subsection{Experimental setup for locomotion experiment}
\label{sec:M_locomotion}

Three modules were connected in a ring and each was equipped with an L-shaped leg, as described in \S\ref{sec:locomotion}.
Each leg was 3D-printed from polylactic acid (PLA), with gripping tape (3M TB641) bonded to the distal end to form the frictional claw.
The leg was joined to the module body with double-sided adhesive through a flexible hinge cut from 0.007-inch (0.18~mm) polyester film (Grafix Dura-Lar).

The robot was placed on a flat acrylic surface and operated at 9.0~V for six complete cycles over a total of 95~minutes.
Displacement was measured after a further 10-minute cooling period, and was tracked from the distal end of the feet on the last module, $\m{3}$, which served as the reference point.
Video was recorded throughout with a mirrorless digital camera (Panasonic LUMIX G9) at 1 frame per second.

\bibliography{reference}

@article{hodgkin_quantitative_1952,
	title        = {A quantitative description of membrane current and its application to conduction and excitation in nerve},
	author       = {Hodgkin, A. L. and Huxley, A. F.},
	volume       = 117,
	pages        = {500--544},
	journal =  {J. Physiol.},
	date         = 1952
}

@article{bean_action_2007,
	title        = {The action potential in mammalian central neurons},
	author       = {Bean, B. P.},
	volume       = 8,
	pages        = {451--465},
	journal =  {Nat. Rev. Neurosci.},
	date         = 2007
}

@article{abbott_lapicques_1999,
	title        = {Lapicque's introduction of the integrate-and-fire model neuron (1907)},
	author       = {Abbott, L. F.},
	volume       = 50,
	number       = 5,
	pages        = {303--304},
	journal =  {Brain Res. Bull.},
	date         = 1999
}

@article{burkitt_review_2006,
	title        = {A review of the integrate-and-fire neuron model: I. Homogeneous synaptic input},
	author       = {Burkitt, A. N.},
	volume       = 95,
	number       = 1,
	pages        = {1--19},
	journal =  {Biol. Cybern.},
	date         = 2006
}

@article{marder_central_2001,
	title        = {Central pattern generators and the control of rhythmic movements},
	author       = {Marder, E. and Bucher, D.},
	volume       = 11,
	number       = 23,
	pages        = {R986--R996},
	journal =  {Curr. Biol.},
	date         = 2001
}

@article{marder_invertebrate_2005,
	title        = {Invertebrate central pattern generation moves along},
	author       = {Marder, E. and Bucher, D. and Schulz, D. J. and Taylor, A. L.},
	volume       = 15,
	number       = 17,
	pages        = {R685--R699},
	journal =  {Curr. Biol.},
	date         = 2005
}

@article{grillner_biological_2006,
	title        = {Biological pattern generation: the cellular and computational logic of networks in motion},
	author       = {Grillner, S.},
	volume       = 52,
	number       = 5,
	pages        = {751--766},
	journal =  {Neuron},
	date         = 2006
}

@article{kristan_neuronal_2005,
	title        = {Neuronal control of leech behavior},
	author       = {Kristan, W. B. and Calabrese, R. L. and Friesen, W. O.},
	volume       = 76,
	number       = 5,
	pages        = {279--327},
	journal =  {Prog. Neurobiol.},
	date         = 2005
}

@article{mulloney_neurobiology_2012,
	title        = {Neurobiology of the crustacean swimmeret system},
	author       = {Mulloney, B. and Smarandache-Wellmann, C.},
	volume       = 96,
	number       = 2,
	pages        = {242--267},
	journal =  {Prog. Neurobiol.},
	date         = 2012
}

@article{ijspeert_central_2008,
	title        = {Central pattern generators for locomotion control in animals and robots: a review},
	author       = {Ijspeert, A. J.},
	volume       = 21,
	number       = 4,
	pages        = {642--653},
	journal =  {Neural Netw.},
	date         = 2008
}

@article{ijspeert_swimming_2007,
	title        = {From swimming to walking with a salamander robot driven by a spinal cord model},
	author       = {Ijspeert, A. J. and Crespi, A. and Ryczko, D. and Cabelguen, J.-M.},
	volume       = 315,
	number       = 5817,
	pages        = {1416--1420},
	journal =  {Science},
	date         = 2007
}

@article{owaki_simple_2013,
	title        = {Simple robot suggests physical interlimb communication is essential for quadruped walking},
	author       = {Owaki, D. and Kano, T. and Nagasawa, K. and Tero, A. and Ishiguro, A.},
	volume       = 10,
	number       = 78,
	pages        = 20120669,
	journal =  {J. R. Soc. Interface},
	date         = 2013
}

@article{chiel_brain_1997,
	title        = {The brain has a body: adaptive behavior emerges from interactions of nervous system, body and environment},
	author       = {Chiel, H. J. and Beer, R. D.},
	volume       = 20,
	number       = 12,
	pages        = {553--557},
	journal =  {Trends Neurosci.},
	date         = 1997
}

@article{hauser_towards_2011,
	title        = {Towards a theoretical foundation for morphological computation with compliant bodies},
	author       = {Hauser, H. and Ijspeert, A. J. and Füchslin, R. M. and Pfeifer, R. and Maass, W.},
	volume       = 105,
	number       = 5,
	pages        = {355--370},
	journal =  {Biol. Cybern.},
	date         = 2011
}

@article{tanaka_recent_2019,
	title        = {Recent advances in physical reservoir computing: a review},
	author       = {Tanaka, G. and Yamane, T. and Héroux, J. B. and Nakane, R. and Kanazawa, N. and Takeda, S. and Numata, H. and Nakano, D. and Hirose, A.},
	volume       = 115,
	pages        = {100--123},
	journal =  {Neural Netw.},
	date         = 2019
}

@article{sitti_physical_2021,
	title        = {Physical intelligence as a new paradigm},
	author       = {Sitti, M.},
	volume       = 46,
	pages        = 101340,
	journal =  {Extreme Mech. Lett.},
	date         = 2021
}

@article{yasuda_mechanical_2021,
	title        = {Mechanical computing},
	author       = {Yasuda, H. and Buskohl, P. R. and Gillman, A. and Murphey, T. D. and Stepney, S. and Vaia, R. A. and Raney, J. R.},
	volume       = 598,
	number       = 7879,
	pages        = {39--48},
	journal =  {Nature},
	date         = 2021
}

@article{yoshida_self-oscillating_1996,
	title        = {Self-oscillating gel},
	author       = {Yoshida, R. and Takahashi, T. and Yamaguchi, T. and Ichijo, H.},
	volume       = 118,
	number       = 21,
	pages        = {5134--5135},
	journal =  {J. Am. Chem. Soc.},
	date         = 1996
}

@article{maeda_peristaltic_2008,
	title        = {Peristaltic motion of polymer gels},
	author       = {Maeda, S. and Hara, Y. and Yoshida, R. and Hashimoto, S.},
	volume       = 47,
	number       = 35,
	pages        = {6690--6693},
	journal =  {Angew. Chem. Int. Ed.},
	date         = 2008
}

@article{yashin_pattern_2006,
	title        = {Pattern formation and shape changes in self-oscillating polymer gels},
	author       = {Yashin, V. V. and Balazs, A. C.},
	volume       = 314,
	number       = 5800,
	pages        = {798--801},
	journal =  {Science},
	date         = 2006
}

@article{raney_stable_2016,
	title        = {Stable propagation of mechanical signals in soft media using stored elastic energy},
	author       = {Raney, J. R. and Nadkarni, N. and Daraio, C. and Kochmann, D. M. and Lewis, J. A. and Bertoldi, K.},
	volume       = 113,
	number       = 35,
	pages        = {9722--9727},
	journal =  {Proc. Natl. Acad. Sci. U.S.A.},
	date         = 2016
}

@article{zhao_soft_2019,
	title        = {Soft phototactic swimmer based on self-sustained hydrogel oscillator},
	author       = {Zhao, Y. and Xuan, C. and Qian, X. and Alsaid, Y. and Hua, M. and Jin, L. and He, X.},
	volume       = 4,
	number       = 33,
	pages        = {eaax7112},
	journal =  {Sci. Robot.},
	date         = 2019
}

@article{proskurkin_experimental_2020,
	title        = {Experimental verification of an opto-chemical “neurocomputer”},
	author       = {Proskurkin, I. S. and Smelov, P. S. and Vanag, V. K.},
	volume       = 22,
	number       = 34,
	pages        = {19359--19367},
	journal =  {Phys. Chem. Chem. Phys.},
	date         = 2020
}

@article{rus_design_2015,
	title        = {Design, fabrication and control of soft robots},
	author       = {Rus, D. and Tolley, M. T.},
	volume       = 521,
	number       = 7553,
	pages        = {467--475},
	journal =  {Nature},
	date         = 2015
}

@article{niiyama_pouch_2015,
	title        = {Pouch motors: printable soft actuators integrated with computational design},
	author       = {Niiyama, R. and Sun, X. and Sung, C. and An, B. and Rus, D. and Kim, S.},
	volume       = 2,
	number       = 2,
	pages        = {59--70},
	journal =  {Soft Robot.},
	date         = 2015
}

@article{miriyev_soft_2017,
	title        = {Soft material for soft actuators},
	author       = {Miriyev, A. and Stack, K. and Lipson, H.},
	volume       = 8,
	pages        = 596,
	journal =  {Nat. Commun.},
	date         = 2017
}

@article{han_untethered_2019,
	title        = {Untethered soft actuators by liquid–vapor phase transition: remote and programmable actuation},
	author       = {Han, J. and Jiang, W. and Niu, D. and {others}},
	volume       = 1,
	number       = 8,
	pages        = 1900109,
	journal =  {Adv. Intell. Syst.},
	date         = 2019
}

@article{mirvakili_actuation_2020,
	title        = {Actuation of untethered pneumatic artificial muscles and soft robots using magnetically induced liquid-to-gas phase transitions},
	author       = {Mirvakili, S. M. and Sim, D. and Hunter, I. W. and Langer, R.},
	volume       = 5,
	number       = 41,
	pages        = {eaaz4239},
	journal =  {Sci. Robot.},
	date         = 2020
}

@article{liu_ethanol_2021,
	title        = {Ethanol phase change actuator based on thermally conductive material for fast cycle actuation},
	author       = {Liu, Z. and Sun, B. and Hu, J. and Zhang, Y. and Lin, Z. and Liang, Y.},
	volume       = 13,
	number       = 23,
	pages        = 4095,
	journal =  {Polymers},
	date         = 2021
}

@article{rothemund_soft_2018,
	title        = {A soft, bistable valve for autonomous control of soft actuators},
	author       = {Rothemund, P. and Ainla, A. and Belding, L. and Preston, D. J. and Kurihara, S. and Suo, Z. and Whitesides, G. M.},
	volume       = 3,
	number       = 16,
	pages        = {eaar7986},
	journal =  {Sci. Robot.},
	date         = 2018
}

@article{ceron_soft_2021,
	title = {Soft Robotic Oscillators With Strain-Based Coordination},
	volume = {6},
	issn = {2377-3766},
	pages = {7557--7563},
	number = {4},
	journal = {{IEEE} Robotics and Automation Letters},
	author = {Ceron, Steven and Kimmel, Marta An and Nilles, Alexandra and Petersen, Kirstin},
	urldate = {2026-08-15},
	date = {2021-10},
}

@article{preston_soft_2019,
	title        = {A soft ring oscillator},
	author       = {Preston, D. J. and Jiang, H. J. and Sanchez, V. and Rothemund, P. and Rawson, J. and Nemitz, M. P. and Lee, W.-K. and Suo, Z. and Walsh, C. J. and Whitesides, G. M.},
	volume       = 4,
	number       = 31,
	pages        = {eaaw5496},
	journal =  {Sci. Robot.},
	date         = 2019
}

@article{preston_digital_2019,
	title        = {Digital logic for soft devices},
	author       = {Preston, D. J. and Rothemund, P. and Jiang, H. J. and Nemitz, M. P. and Rawson, J. and Suo, Z. and Whitesides, G. M.},
	volume       = 116,
	number       = 16,
	pages        = {7750--7759},
	journal =  {Proc. Natl. Acad. Sci. U.S.A.},
	date         = 2019
}

@article{drotman_electronics-free_2021,
	title        = {Electronics-free pneumatic circuits for controlling soft-legged robots},
	author       = {Drotman, D. and Jadhav, S. and Sharp, D. and Chan, C. and Tolley, M. T.},
	volume       = 6,
	number       = 51,
	pages        = {eaay2627},
	journal =  {Sci. Robot.},
	date         = 2021
}

@article{laake_fluidic_2022,
	title        = {A fluidic relaxation oscillator for reprogrammable sequential actuation in soft robots},
	author       = {Laake, L. C. van and Vries, J. de and Kani, S. Malek and Overvelde, J. T. B.},
	volume       = 5,
	number       = 9,
	pages        = {2898--2917},
	journal =  {Matter},
	date         = 2022
}

@inproceedings{nemitz_soft_2020,
	title        = {Soft non-volatile memory for non-electronic information storage in soft robots},
	author       = {Nemitz, M. P. and Abrahamsson, C. K. and Wille, L. and Stokes, A. A. and Preston, D. J. and Whitesides, G. M.},
	booktitle    = {2020 {IEEE} Int. Conf. Soft Robotics ({RoboSoft})},
	pages        = {7--12},
	date         = 2020
}

@article{conrad_3d-printed_2024,
	title        = {3D-printed digital pneumatic logic for the control of soft robotic actuators},
	author       = {Conrad, S. and Teichmann, J. and Auth, P. and Knorr, N. and Ulrich, K. and Bellin, D. and Speck, T. and Tauber, F. J.},
	volume       = 9,
	number       = 86,
	pages        = {eadh4060},
	journal =  {Sci. Robot.},
	date         = 2024
}

@article{picella_pneumatic_2024,
	title        = {Pneumatic coding blocks enable programmability of electronics-free fluidic soft robots},
	author       = {Picella, S. and Riet, C. M. van and Overvelde, J. T. B.},
	volume       = 10,
	number       = 51,
	pages        = {eadr2433},
	journal =  {Sci. Adv.},
	date         = 2024
}

@article{shepherd_multigait_2011,
	title        = {Multigait soft robot},
	author       = {Shepherd, R. F. and Ilievski, F. and Choi, W. and Morin, S. A. and Stokes, A. A. and Mazzeo, A. D. and Chen, X. and Wang, M. and Whitesides, G. M.},
	volume       = 108,
	number       = 51,
	pages        = {20400--20403},
	journal =  {Proc. Natl. Acad. Sci. U.S.A.},
	date         = 2011
}

@article{tolley_resilient_2014,
	title        = {A resilient, untethered soft robot},
	author       = {Tolley, M. T. and Shepherd, R. F. and Mosadegh, B. and Galloway, K. C. and Wehner, M. and Karpelson, M. and Wood, R. J. and Whitesides, G. M.},
	volume       = 1,
	number       = 3,
	pages        = {213--223},
	journal =  {Soft Robot.},
	date         = 2014
}

@article{comoretto_physical_2025,
	title        = {Physical synchronization of soft self-oscillating limbs for fast and autonomous locomotion},
	author       = {Comoretto, A. and Schomaker, H. A. H. and Overvelde, J. T. B.},
	volume       = 388,
	number       = 6647,
	pages        = {610--615},
	journal =  {Science},
	date         = 2025
}

@article{quillin_kinematic_1999,
	title        = {Kinematic scaling of locomotion by hydrostatic animals: ontogeny of peristaltic crawling by the earthworm ıt Lumbricus terrestris},
	author       = {Quillin, K. J.},
	volume       = 202,
	number       = 6,
	pages        = {661--674},
	journal =  {J. Exp. Biol.},
	date         = 1999
}

@article{bassler_pattern_1998,
	title        = {Pattern generation for stick insect walking movements—multisensory control of a locomotor program},
	author       = {Bässler, Ulrich and Büschges, Ansgar},
	volume       = 27,
	number       = 1,
	pages        = {65--88},
	journal =  {Brain Research Reviews},
	shortjournal = {Brain Research Reviews},
	date         = {1998-06-01}
}

@article{gockowski_improving_2025,
  title   = {Improving the efficiency of soft phase-change actuators using thermodynamic analysis},
  author  = {Gockowski, L. F. and Xiao, C. and Hao, A. and Zhu, Y. and Liao, B. and Valentine, M. T. and Hawkes, E. W.},
  journal = {Soft Robot.},
  volume  = {12},
  pages   = {687},
  year    = {2025},
  doi     = {10.1089/soro.2024.0139}
}

\bibliographystyle{sciencemag}

\section*{Acknowledgments}

\paragraph*{Funding:}
This work was supported by the Office of Naval Research (ONR) under grant numbers N00014-23-1-2358 (N.G., M.T. and D.L.), N00014-23-1-2169 (N.G., M.T. and D.L.) and N00014-26-1-2132 (N.G. and D.L.). Any opinions, findings, conclusions, or recommendations expressed in this material are those of the authors and do not necessarily reflect the views of the ONR.

\paragraph*{Author contributions:}
Conceptualization: D.L., M.T.T., and N.G. Methodology: D.L., M.T.T., and N.G. Validation: D.L. Formal analysis: D.L. Investigation: D.L. Resources: M.T.T. and N.G. Data curation: D.L. Writing---original draft: D.L. Writing---review and editing: D.L., M.T.T., and N.G. Visualization: D.L. Supervision: M.T.T. and N.G. Project administration: M.T.T. and N.G. Funding acquisition: M.T.T. and N.G.

\paragraph*{Competing interests:}

The authors declare that they have no competing interests.

\paragraph*{Data and materials availability:}
All study data are included in the article and/or supporting information. All materials associated with this study are listed in the Methods section and/or are available commercially

\newpage

\end{document}